\documentclass{article}

\usepackage[preprint,nonatbib]{neurips_2023}

\usepackage[utf8]{inputenc} 
\usepackage[T1]{fontenc}    
\usepackage{hyperref}       
\usepackage{url}            
\usepackage{booktabs}       
\usepackage{amsfonts}       
\usepackage{nicefrac}       
\usepackage{graphicx}
\usepackage{subcaption}

\usepackage{microtype}      
\usepackage{xcolor}         
\usepackage{algorithm}
\usepackage{algorithmic}
\usepackage{amsmath}

\title{Latent Painter}
\newcommand{\argmax}{\arg\!\max}

\author{%
  Shih-Chieh Su\\
  \texttt{jessysu@gmail.com} \\
}

\begin{document}

\maketitle

\begin{abstract}
  Latent diffusers revolutionized the generative AI and inspired creative art. When denoising the latent, the predicted original image at each step collectively animates the formation. However, the animation is limited by the denoising nature of the diffuser, and only renders a sharpening process. This work presents Latent Painter, which uses the latent as the canvas, and the diffuser predictions as the plan, to generate painting animation. Latent Painter also transits one generated image to another, which can happen between images from two different sets of checkpoints.
\end{abstract}

\section{Introduction}

Can you imagine the generated picture having its own painting action? Or transit between multiple images generated by different pre-trained diffusers? Enabling such function extends the capability of the denoising diffusion, and the proposed method works on any existing diffuser.

Recently, denoising diffusers gain a lot of traction in generative AI, for its high quality outcome without adversarial training~\cite{ho2020denoising}, its efficiency~\cite{song2020denoising, karras2022elucidating}, content diversity with easy text conditioning~\cite{ramesh2022hierarchical}, and reasonable footprint~\cite{rombach2022high}. Although the convenience of text-to-image largely spurs the creativity, little has be studied about the composition of its generated art. This work presents Latent Painter, which uses the existing diffuser to generate painting animation along with the output image. 

During the diffusion denoising, the latent is denoised step-by-step into the state representing an image matching the text input. The predicted original image, which is the progressive estimate $\hat{x_0}$ of the reverse diffusion process in ~\cite{ho2020denoising}, becomes sharper and sharper when the latent being denoised. Collecting the predicted original images forms an animation about the sharpening process, where information is updated omnipresently in the same frame. However, the update is uneven across frames, with higher total pixel value change toward earlier frames. Latent painter prioritizes the update locally to just like the brush strokes. Once the released information is close enough to match the current predicted original, the residue is accumulated for later updates. This mechanism provides update more evenly over frames.

\section{Method}

Let $Z(x,y,c)$ denote the state of the latent, having shape of width by height by channels, and can be initiated as zeros. During each inference step $t$, the scheduler provides an updated denoised latent sample $D_t$. The policy $\mathcal{P}$ decides whether the information difference between $Z$ and $D_t$ qualifies an update and if so, which subset ($\bar{C}$) of all latent channels $C$ needs to be updated. 
\begin{equation} \label{eq:policy}
\bar{C} = \mathcal{P}(Z, D_t)\text{, where }\bar{C} \in C
\end{equation}
During a channel update, a channel $c$ is chosen out of $\bar{C}$. Then, a qualifying threshold $\theta$ picks the region $R$ to be stroked on.
\begin{equation} \label{eq:qaul_region}
R = \{(x,y): |Z(x,y,c)-D_t(x,y,c)|>\theta\}
\end{equation}

Let $G_c(x,y)$ denote the information gain at the current channel $c$.
\begin{equation} \label{eq:info_gain}
G_c(x,y) = |Z_c(x,y) - D_{t,c}(x,y)|
\end{equation}
The first stroke is placed at the location whose neighborhood $\mathcal{N}(x,y)$ has the largest information gain.
\begin{equation} \label{eq:stroke_point}
p(\hat{x},\hat{y}) = \argmax_{(x,y)} \sum_{(x',y')\in\mathcal{N}(x,y)} G_c(x',y')
\end{equation}
The whole neighborhood $\mathcal{N}(x,y)$, presented as the stroke of the Latent Painter, is then updated with the current scheduler output.
\begin{equation} \label{eq:stroke_update}
Z(x,y,c)=D_t(x,y,c) \text{, for all } (x,y)\in\mathcal{N}(\hat{x},\hat{y})
\end{equation}
The stroke-able region in Eq.~\ref{eq:qaul_region} is then updated to exclude the newly stroke region. Following the same procedure, the strokes are placed one by one until $R$ being empty, or an early stopping criteria $\mathcal{E}$ has been reached. Continue the stroke action in other channels in $\bar{C}$ likewise.

Upon finishing all $\bar{C}$ channels, the painter steps through another iteration $t$ of the scheduler to get a new $D_t$, then starts from Eq.~\ref{eq:policy} to get the channels to be painted this iteration. However, there could be very little difference between the new $D_t$ and current $Z$. This is particularly true toward the end of the denoising process. 

To ensure the animation frames having meaningful update, one idea is to release $D_t$ more evenly across strokes, each being presented in one frame. Therefore, $\mathcal{P}$ requires the total difference between $Z$ and $D_t$ being larger than a portion of the largest total difference over all previous iterations. With $D_t$ sufficiently diverse from $Z$, $\mathcal{P}$ provides the channels needing update and to be painted, as in Eq.~\ref{eq:policy}. 

Note the current scheduler output $D_t$ may not fully pass onto the latent $Z$ at time $t$, when the early stopping condition $\mathcal{E}$ exists. The un-updated residue is carried over to next scheduler output $D_{t+1}$. The accumulated residue becomes the momentum for the next strokes.

\subsection{Cost}

\begin{algorithm}   
    \caption{Latent Painter - Strokes \\
    Given policy $\mathcal{P}$, stroke qualifier $\theta$, stroke neighborhood $\mathcal{N}$ } \label{alg:stroke}
    \begin{algorithmic}
        \STATE Initialize painter state $Z$ as zeros of the latent shape $w\times h\times |C|$
        \FOR{$t$ in diffuser schedule}
        \STATE
        Compute $D_t$, the diffuser latent space prediction of $x_0$ at current time $t$
        \STATE
        Based on policy $\mathcal{P}$, decides the channels to be updated $\bar{C} = \mathcal{P}(Z, D_t)$, where $\bar{C} \in C$
        \FOR{$c \in \bar{C} $}
        \STATE Compute the stroke region $R = \{(x,y): |Z(x,y,c)-D_t(x,y,c)|>\theta\}$
        \STATE Initialize move cost $M(x,y)$ as $w\times h$ of ones
        \WHILE{$R$ not empty, as step $s$, }
        \STATE  
        Compute gain $G_c(x,y) = |Z_c(x,y) - D_{t,c}(x,y)|$ at channel $c$
        \STATE Compute motivation $V(x,y) = G_c(x,y) \cdot M_c(x,y)$ 
        \STATE Pick the stroke point $p(\hat{x},\hat{y}) = \argmax_{(x,y)} \sum_{(x',y')\in\mathcal{N}(x,y)} V(x',y')$
        \STATE Stroke to make $Z(x,y,c)=D_t(x,y,c)$ for all $(x,y)\in\mathcal{N}(\hat{x},\hat{y})$
        \STATE Update $R \leftarrow R $ \textbackslash $\{(x,y) : (x,y) \in\mathcal{N}(\hat{x},\hat{y})\}$
        \STATE Compute move cost $M(x,y)$ as an inverse Gaussian filter centered $(\hat{x},\hat{y})$
        \ENDWHILE
        \ENDFOR
        \ENDFOR
    \end{algorithmic}
\end{algorithm}

The human painter typically considers optimizing the effort in painting, such as to stroke the nearby area first, and keep using the same brush and the same color as much as possible, before changing or cleaning the brush. As for machine, while the convolutional layers are trained to capture congruent patterns, placing the strokes within the same channel allows more congruent patterns being stroked.

The moving cost of the brush, denoted by $M_c(x,y)$, is modeled as the inverse of a Gaussian kernel centered at the current stroke location. Rather than the location with largest information gain, the stroke is placed where having the largest motivation $V$, which is defined as the information gain modulated by the moving cost,
\begin{equation} \label{eq:stroke_motivation}
V(x,y) = G_c(x,y) \cdot M_c(x,y)
\end{equation}

The motivation-based stroke method is presented in Alg.~\ref{alg:stroke}.

\section{Result}

\begin{figure}[ht]
    \centering
    \begin{subfigure}[b]{\textwidth}
        \includegraphics[width=\textwidth]{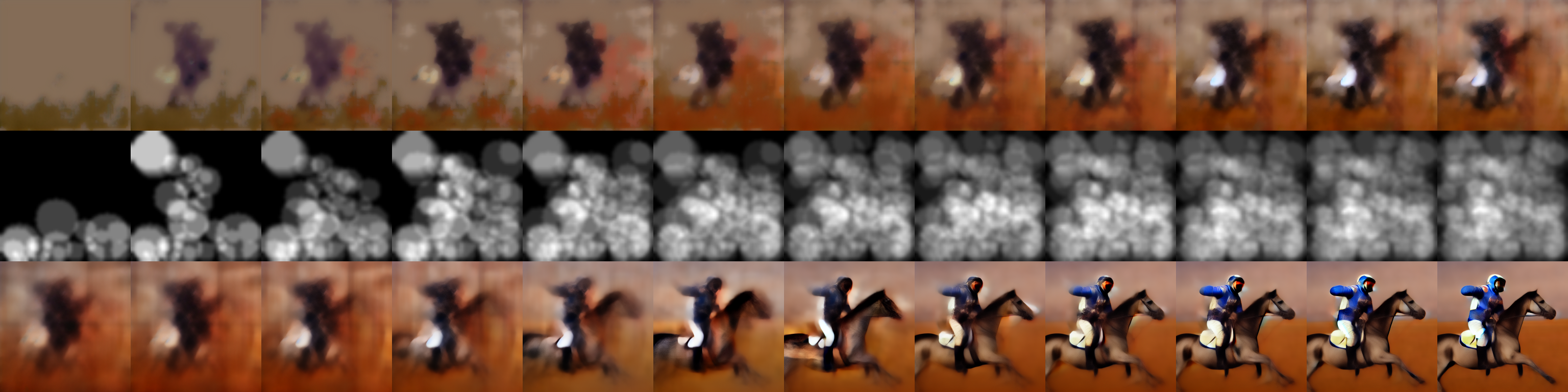}
    \end{subfigure}
    \begin{subfigure}[b]{\textwidth}
        \centering
        \includegraphics[width=\textwidth]{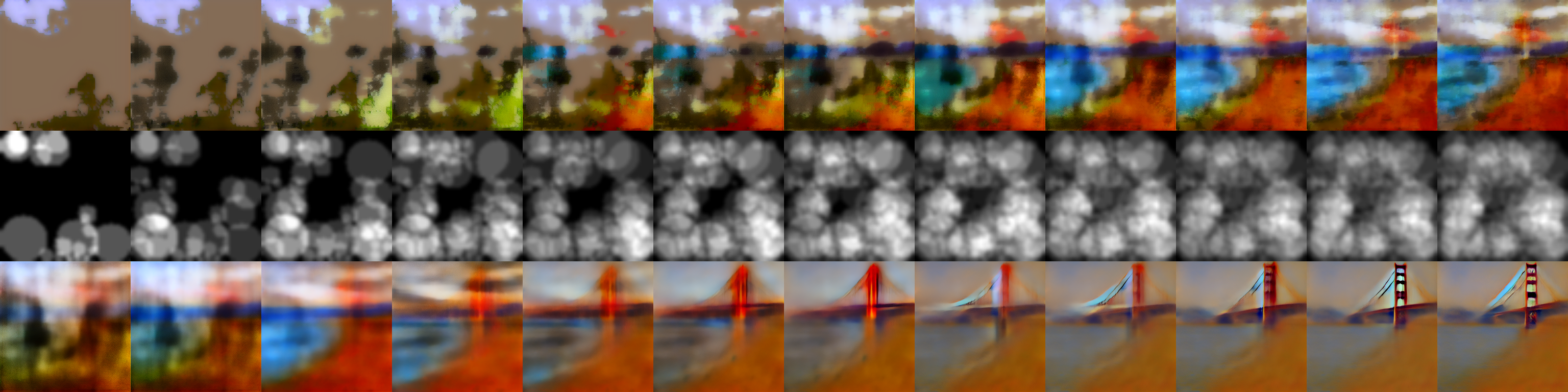}
    \end{subfigure}
    \begin{subfigure}[b]{\textwidth}
        \centering
        \includegraphics[width=\textwidth]{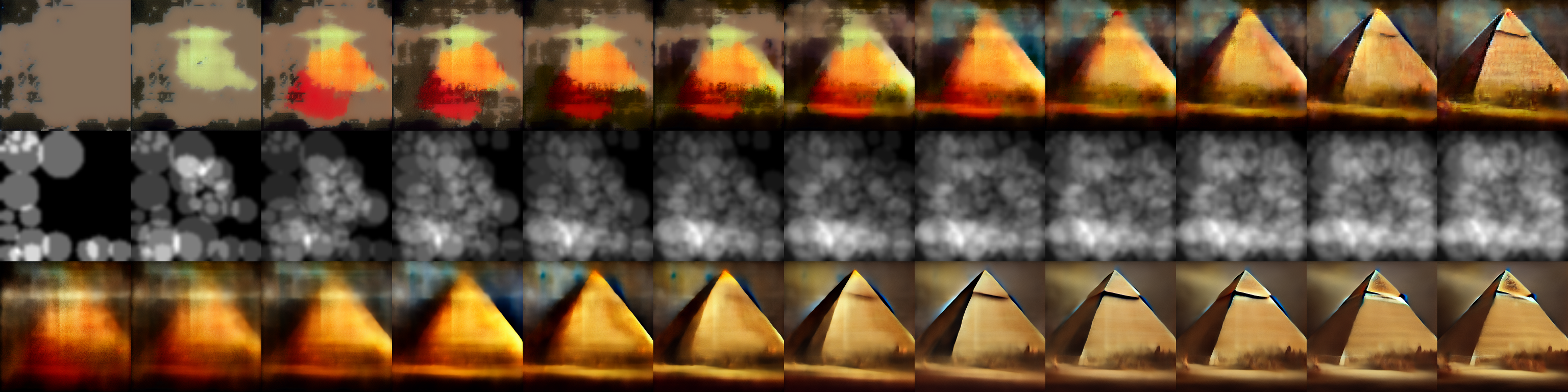}
    \end{subfigure}
    \begin{subfigure}[b]{\textwidth}
        \centering
        \includegraphics[width=\textwidth]{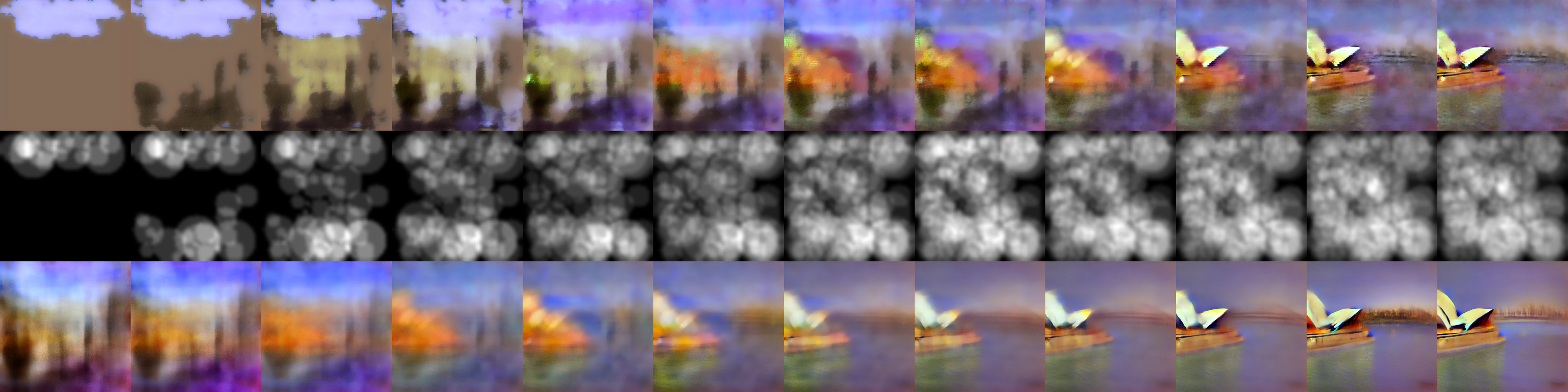}
    \end{subfigure}
    \caption{Stroke location of Latent Painter. Within each block, the bottom row shows the predicted original images $\hat{x_0}$ after each of the first 12 denoising iterations, each with only one frame. The top row shows the Latent Painter stroke snapshots, each containing 20-40 strokes (frames), during the same period. The middle row are the corresponding accumulated stroke map, stacked over channels. The animations are available at \href{https://latentpainter.github.io/}{https://latentpainter.github.io/}}
    \label{fig:LP_output}
\end{figure}

Sample outputs from the Latent Painter Strokes (Alg.~\ref{alg:stroke}) are shown in Fig.~\ref{fig:LP_output}. The samples were first generated with stable diffusion~\cite{rombach2022high} using a text sentence. The incurred latent series of the predicted original images is fed into the painter to produce the strokes. From the bottom rows of each block in Fig.~\ref{fig:LP_output}, the denoising outputs quickly converge close to the final state in the first couple iterations, each of which provides only one frame in the animation. 

In contrast, the Latent Painter slows down the rapid update during early denoising iterations. This prevents frames being updated too quickly, and helps rendering the new information more evenly over frames. Here, the new information is released in the form of strokes. Each denoising iteration can be released in tens to hundreds of strokes (frames). The stroke maps in the middle rows indicate the regions with large information gap between $D_T$ and $Z$, thus being stroked. 

The brighter regions in the stroke map are from the accumulated strokes in those regions, aggregated across channels. Since the Latent Painter Strokes is driven by the differential content between denoising updates, the stroke map provides a way to observe the informative regions that require intensive updates to yield the final detail.

\subsection{Stroke Content}

\begin{figure}[ht]
    \centering
    \begin{subfigure}[b]{\textwidth}
        \includegraphics[width=0.198\textwidth]{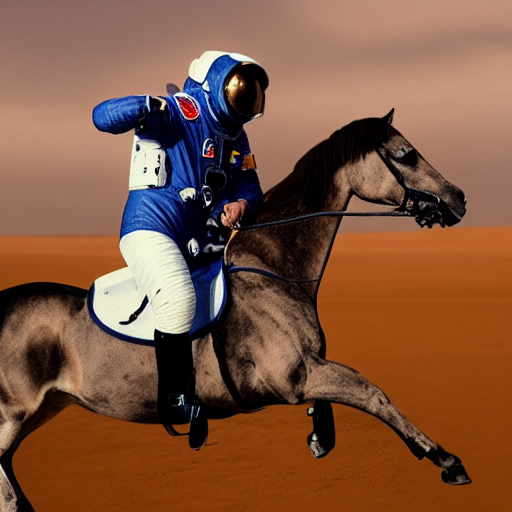}
        \includegraphics[width=0.792\textwidth]{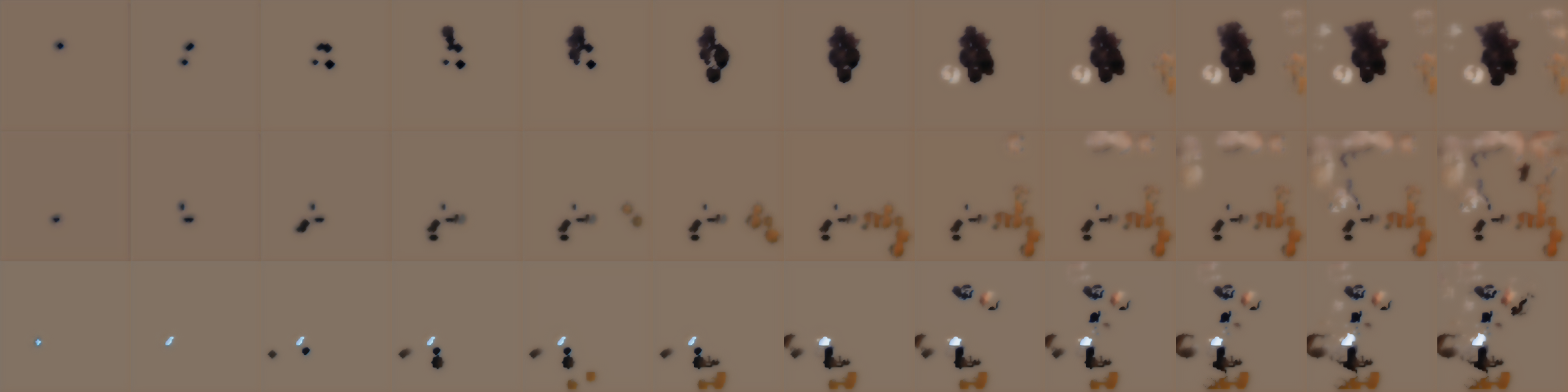}
    \end{subfigure}
    \begin{subfigure}[b]{\textwidth}
        \includegraphics[width=0.198\textwidth]{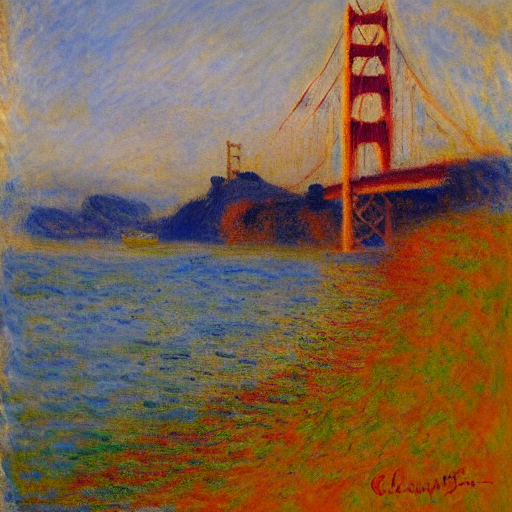}
        \includegraphics[width=0.792\textwidth]{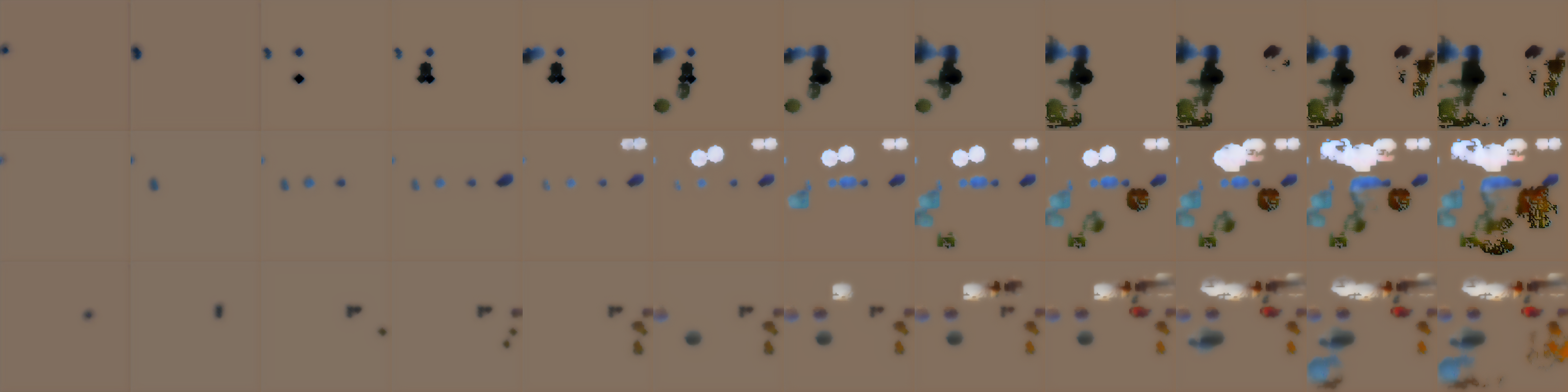}
    \end{subfigure}
    \begin{subfigure}[b]{\textwidth}
        \includegraphics[width=0.198\textwidth]{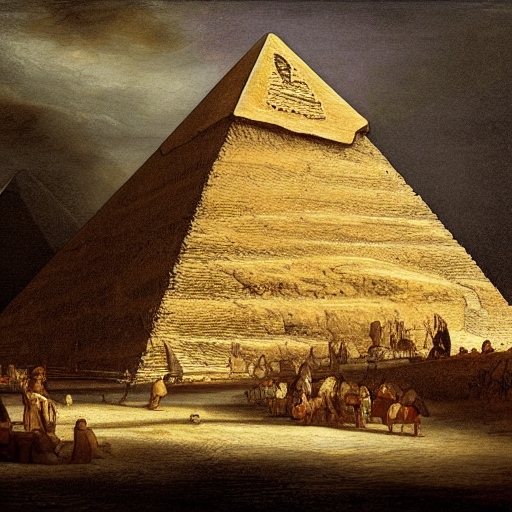}
        \includegraphics[width=0.792\textwidth]{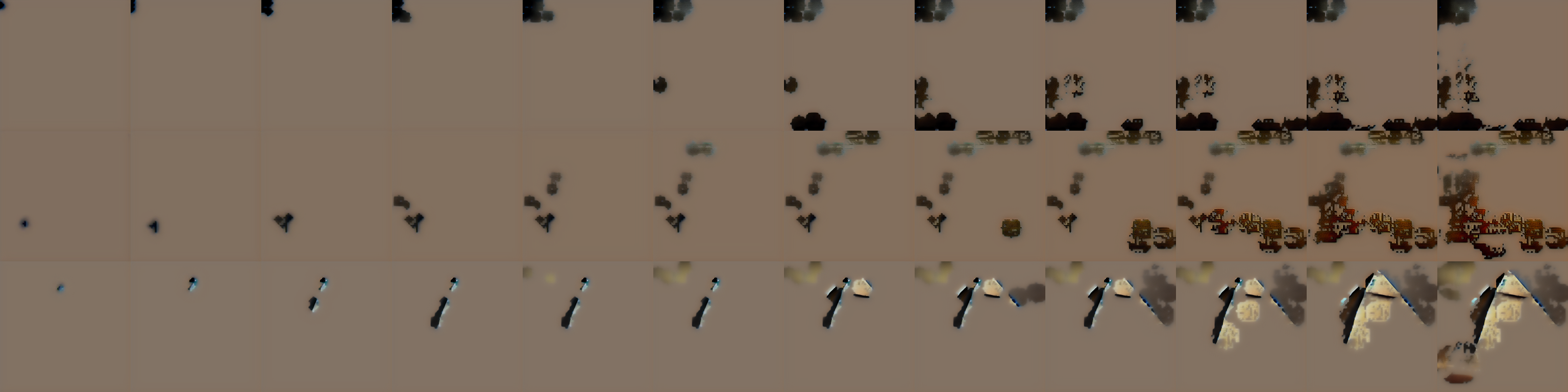}
    \end{subfigure}
    \begin{subfigure}[b]{\textwidth}
        \includegraphics[width=0.198\textwidth]{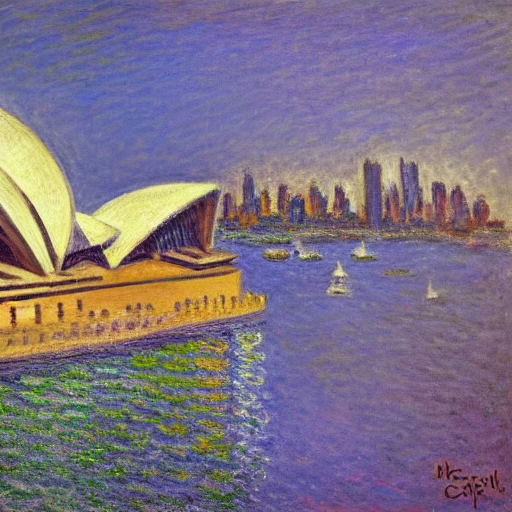}
        \includegraphics[width=0.792\textwidth]{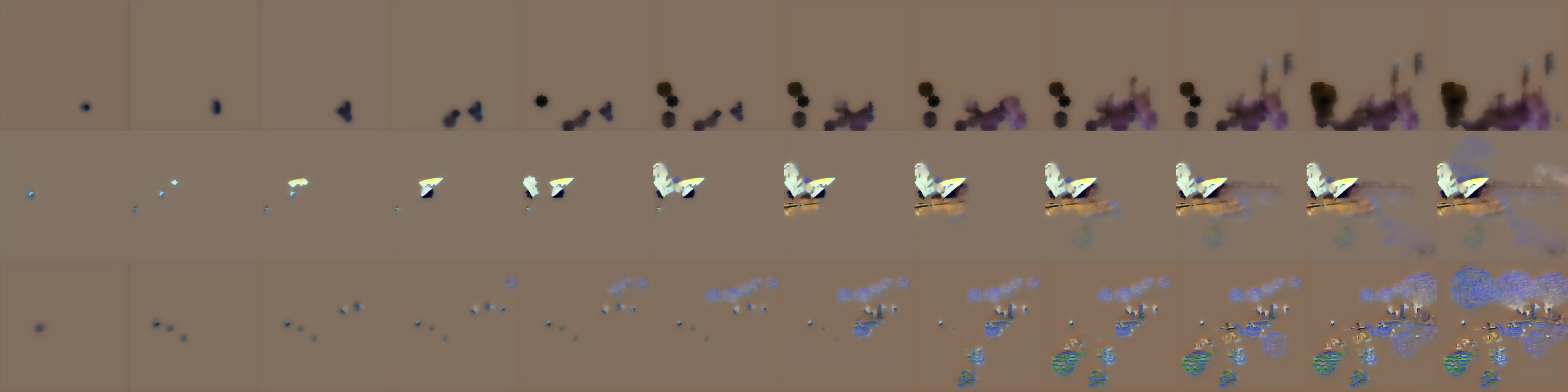}
    \end{subfigure}
    \caption{Stroke content of Latent Painter. Within each block, the rows show the cumulative stroke output of channel 0, when it was chosen out of $\bar{C}$ for the first three times as the target channel to stroke.}
    \label{fig:LP_color}
\end{figure}

What have been stroked onto the canvas? Do the spatially and temporally nearby strokes share similar pattern, color or style? Some stroke samples from the Latent Painter are shown in Fig.~\ref{fig:LP_color}. The snapshots each row only accumulates the strokes from the same channel update, which starts from Eq.~\ref{eq:qaul_region}. Within the same channel, the strokes tend to be congruent in color or style, or both. The channels of the latent space are different from those at the visible layer, where the RGB channels each only account for a color. 

Ultimately, the cardinality in color and style is bounded by the number of trainable neurons, i.e. network parameters. Latent Painter paints at the latent space, which has only four channels per the diffusion system in \cite{rombach2022high}. Designed to be a compact representation space, the latent space has only four channels. Thus even within the same channel, the strokes can contain different styles and colors. However, the denoising process provides the guidance to differentiate the further detail within the channel. Therefore, one can attain better congruence via either lifting the early stop condition $\mathcal{E}$ or ensuring $Z(x,y,c)=D_t(x,y,c)$ for all $(x,y)$ at the end of the update of current $c$.

\subsection{Beyond Strokes}

\begin{figure}[ht]
    \centering
    \begin{subfigure}[b]{\textwidth}
        \includegraphics[width=\textwidth]{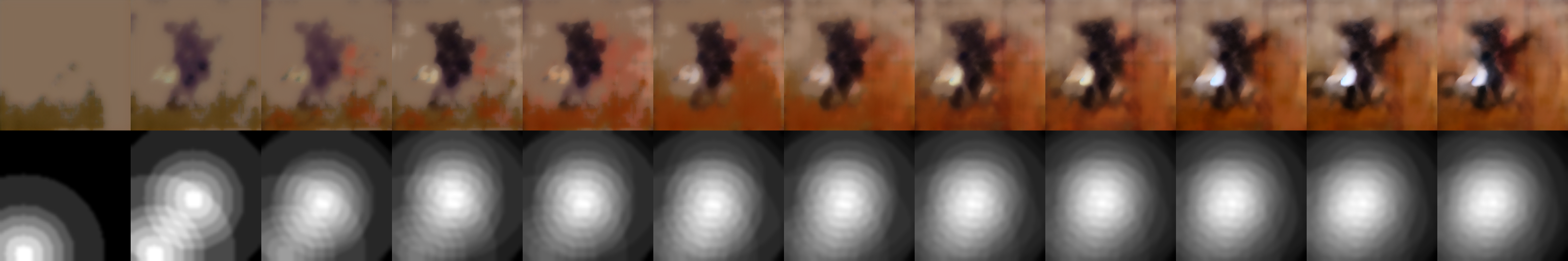}
        \caption{glow}
    \end{subfigure}
    \begin{subfigure}[b]{\textwidth}
        \centering
        \includegraphics[width=\textwidth]{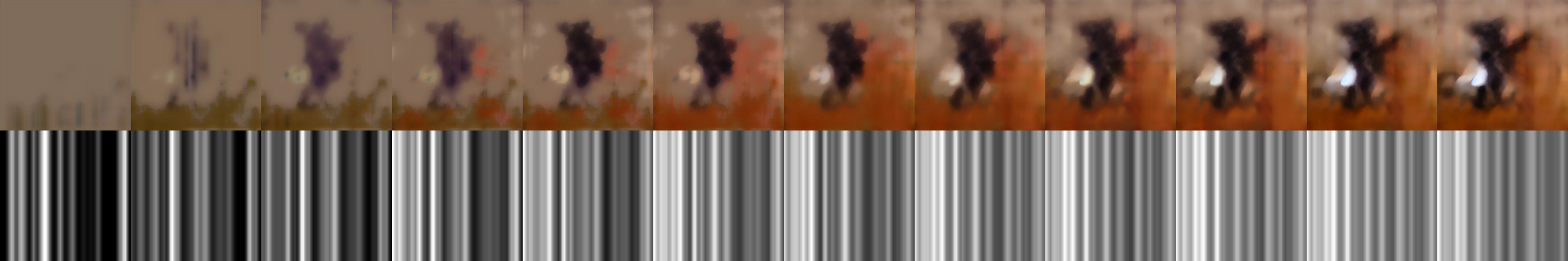}
        \caption{dissolve vertical}
    \end{subfigure}
    \begin{subfigure}[b]{\textwidth}
        \centering
        \includegraphics[width=\textwidth]{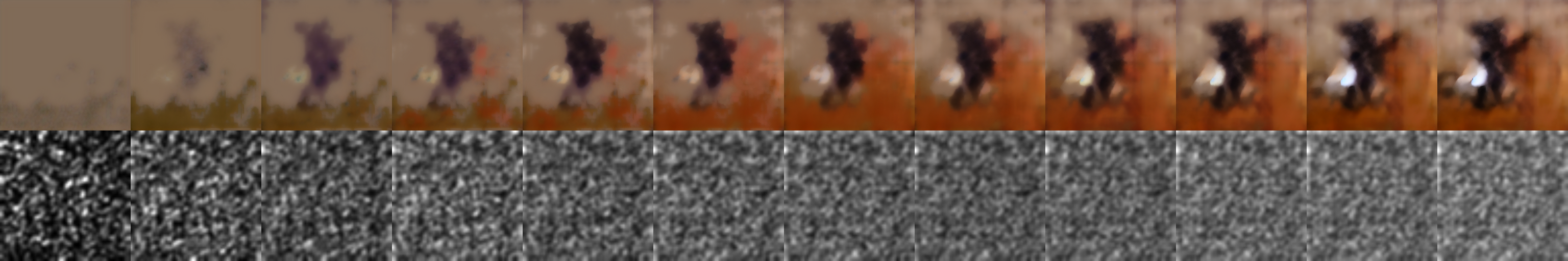}
        \caption{dissolve}
    \end{subfigure}
    \begin{subfigure}[b]{\textwidth}
        \centering
        \includegraphics[width=\textwidth]{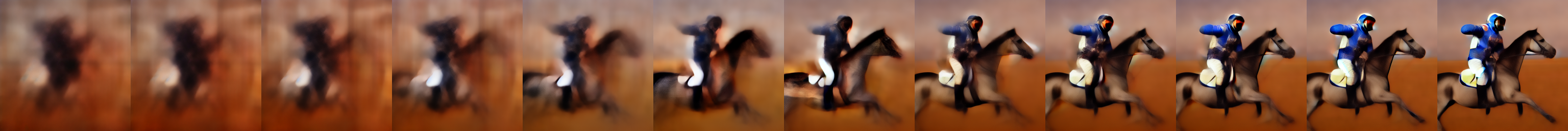}
        \caption{predicted original}
    \end{subfigure}
    \caption{Extensions from strokes. The glow effect is content-driven, while the dissolve here isn't. See \href{https://latentpainter.github.io/}{https://latentpainter.github.io/} for animations and more examples.}
    \label{fig:LP_beyond}
\end{figure}

Besides the painting action in Alg.~\ref{alg:stroke}, there are other ways to leverage the differential response between updates. The glow effect collects updates of the differential latent into the mass center, where the update radiates concentrically.

In addition to the content-driven fashion, Latent Painter can also paint regardless of the differential latent response. Some use cases including the flip effect that mimics page flipping and the fade effect that release the update uniformly. The dissolve effect, however, can be either content-driven or random. Some sample trails of the mentioned effects are visualized in Fig.~\ref{fig:LP_beyond}.

\subsection{Image Transition}

\begin{figure}[ht]
    \centering
    \begin{subfigure}[b]{\textwidth}
        \includegraphics[width=\textwidth]{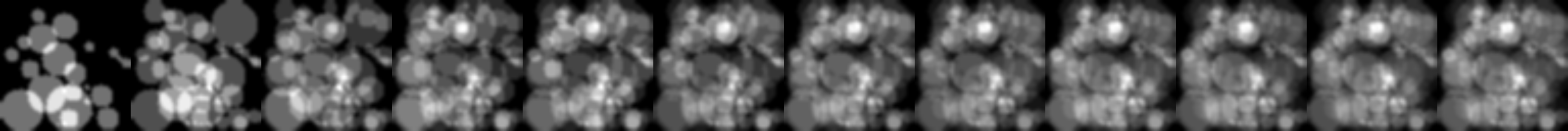}
        \includegraphics[width=\textwidth]{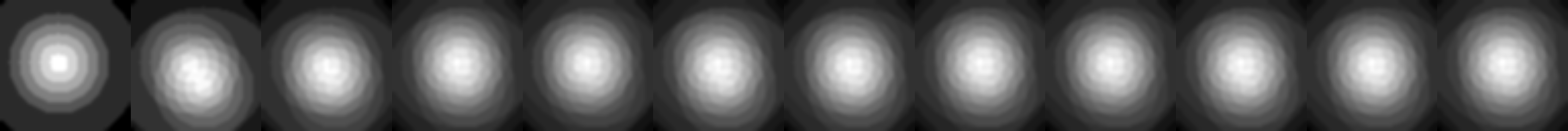}
        \includegraphics[width=\textwidth]{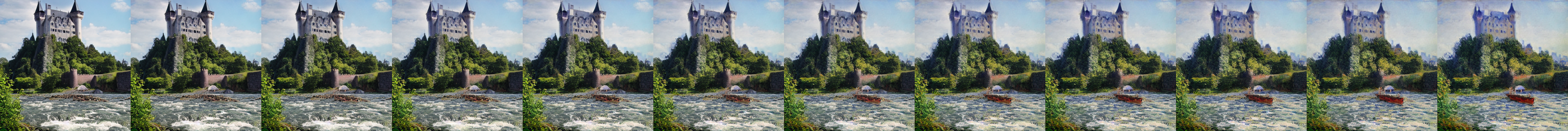}
    \end{subfigure}
    \par\medskip
    \begin{subfigure}[b]{\textwidth}
        \includegraphics[width=\textwidth]{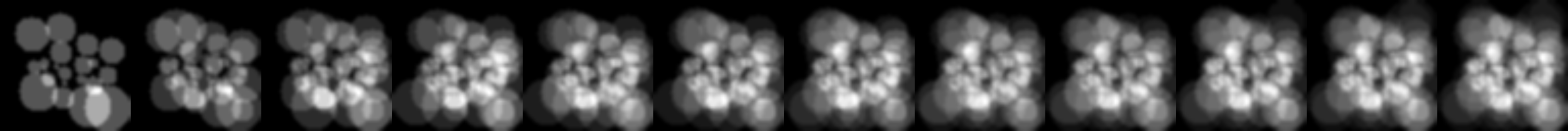}
        \includegraphics[width=\textwidth]{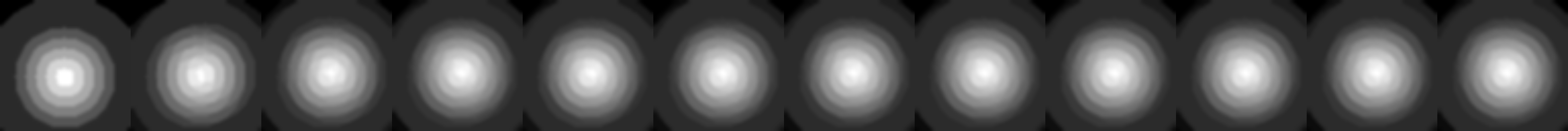}
        \includegraphics[width=\textwidth]{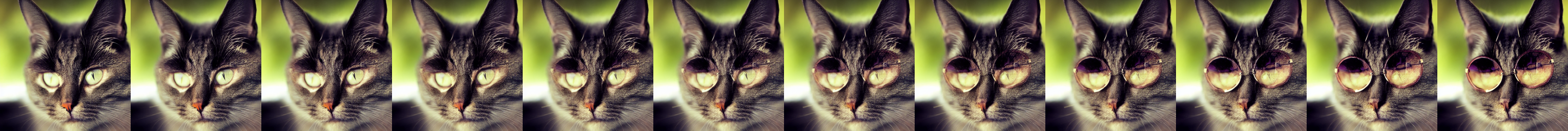}
    \end{subfigure}
    \par\medskip
    \begin{subfigure}[b]{\textwidth}
        \includegraphics[width=\textwidth]{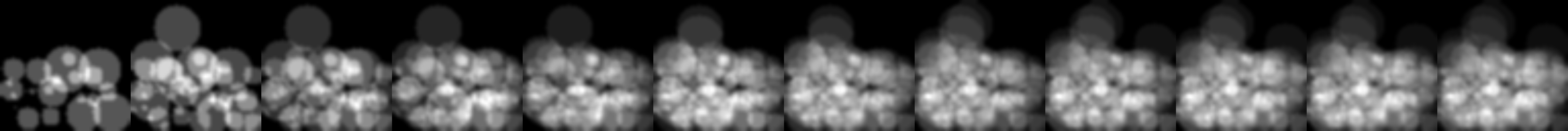}
        \includegraphics[width=\textwidth]{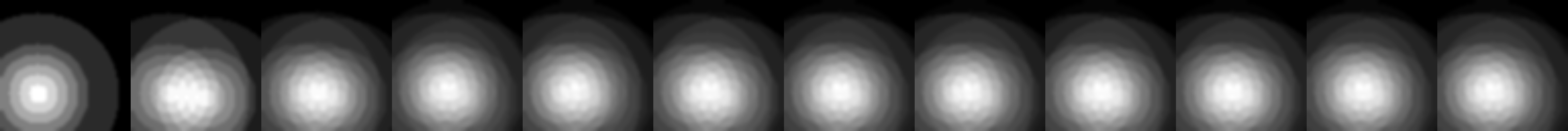}
        \includegraphics[width=\textwidth]{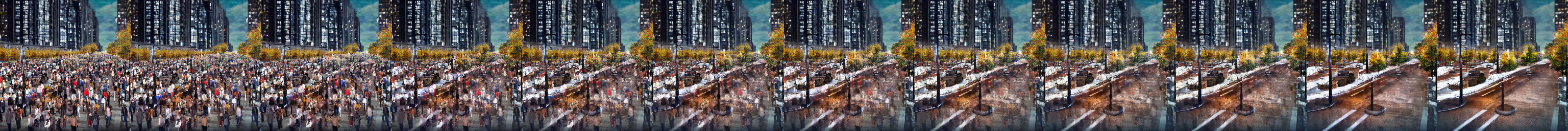}
    \end{subfigure}
    \caption{Image to image transiting animation guided by the interpolated latents from ~\cite{brack2023sega}. Within each block, the top row presents the stroke heatmap transiting one image to its semantically edited variation, while the middle row presents the update heatmap of glow effect, both cumulative. The bottom row shows the output status at the corresponding time points.}
    \label{fig:LP_im2im}
\end{figure}

Existing image transiting animation is based on interpolating the seeded embeddings between the source and the destination, or interpolating the latents, or both. However, it takes either more memory or more time to denoise the interpolated embeddings. Latent Painter animates based on the predicted original images from only two denoising trails, the source and the destination. With the chosen painting effect, the update runs the source image denoising schedule backward, and then the destination schedule forward. Through the constraints of the update release, the transition time can be traded with the detail.

When the source and destination images share a certain part of background, such as in image editing, the interpolated latents between the source latent and the destination latent can be used as the prediction guidance. This avoids using the denoising schedules to guide. Fig~\ref{fig:LP_im2im} illustrates the painting progress to transit the semantically edited images from ~\cite{brack2023sega}.

As an extra credit, Latent Painter can transit the generated images from two different sets of denoising checkpoints, given the same VAE is used to decode the latent.

\section{Conclusion}

Latent Painter is presented in this work. It turns existing latent diffuser output into painting actions via evenly releasing the information update during the denoising iterations. Several extension beyond the stroke algorithm has been covered. Since the latent space of the stable diffusion~\cite{rombach2022high} has only four channels, it is possible to investigate the VAE for further painting behaviors.


\bibliographystyle{unsrt}
\bibliography{ref}

\end{document}